\documentclass[10pt,letterpaper]{article}
\usepackage[top=0.85in,left=2.75in,footskip=0.75in]{geometry}

\usepackage{amsmath,amssymb}

\usepackage{changepage}

\usepackage{textcomp,marvosym}

\usepackage{cite}

\usepackage{nameref,hyperref}

\usepackage[nopatch=eqnum]{microtype}
\DisableLigatures[f]{encoding = *, family = * }

\usepackage[table]{xcolor}

\usepackage{array}

\newcolumntype{+}{!{\vrule width 2pt}}

\newlength\savedwidth

\raggedright
\usepackage[aboveskip=1pt,labelfont=bf,labelsep=period,justification=raggedright,singlelinecheck=off]{caption}

\makeatletter
\renewcommand{\@biblabel}[1]{\quad#1.}
\makeatother

\usepackage{lastpage,fancyhdr,graphicx}
\usepackage{epstopdf}
\fancyheadoffset[L]{2.25in}
\fancyfootoffset[L]{2.25in}
\newcommand{\ANBlock}[1][]{\mathcal{B}_{\mathrm{AN}}}

\begin{document}
\vspace*{0.2in}

\begin{flushleft}
{\Large
\textbf{A neuromorphic model of the insect visual system for natural image processing} %
}
\newline
\\
Adam D. Hines\textsuperscript{1,2}*,
Karin Nordström\textsuperscript{3},
Andrew B. Barron\textsuperscript{1}
\\
\bigskip
\textbf{1} School of Natural Sciences, Macquarie University, Sydney 2109, NSW, Australia
\\
\textbf{2} QUT Centre for Robotics, Queensland University of Technology, Brisbane 4000, QLD, Australia
\\
\textbf{3} College of Medicine and Public Health, Flinders University, Adelaide 5042, SA, Australia
\\
\bigskip

* Corresponding author: adam.hines@mq.edu.au

\end{flushleft}
\section*{Abstract}
Insect vision supports complex behaviors including associative learning, navigation, and object detection, and has long motivated computational models for understanding biological visual processing. However, many contemporary models prioritize task performance while neglecting biologically grounded processing pathways. Here, we introduce a bio-inspired vision model that captures  principles of the insect visual system to transform dense visual input into sparse, discriminative codes. The model is trained using a fully self-supervised contrastive objective, enabling representation learning without labeled data and supporting reuse across tasks without reliance on domain-specific classifiers. We evaluated the resulting representations on flower recognition tasks and natural image benchmarks. The model consistently produced reliable sparse codes that distinguish visually similar inputs. To support different modelling and deployment uses, we have implemented the model as both an artificial neural network and a spiking neural network. In a simulated localization setting, our approach outperformed a simple image downsampling comparison baseline, highlighting the functional benefit of incorporating neuromorphic visual processing pathways. Collectively, these results advance insect computational modelling by providing a generalizable bio-inspired vision model capable of sparse computation across diverse tasks. 

\section*{Author summary}
Insects solve visual problems with tiny brains, so their visual systems offer useful clues for building efficient computer vision. Many insect-inspired vision models, though, are tailored for a single task and don’t transfer well to new problems. Here we present a more general insect-inspired vision model that can be used across multiple tasks without redesigning the system each time. The model processes images through a sequence of stages that mirrors key steps in insect vision and produces a sparse “barcode-like” pattern of activity, similar to how Kenyon cells represent information in insect brains. This compact code can then drive different behaviors or decisions.

\section*{Introduction}

Animal eyes sample more information than animal brains can process.  This statement applies across mammals, reptiles, birds, fish and insects.  A key function of the biological visual system is, therefore, to downsample the information gathered by the retina to a number of sparse neural activations that can be managed by the central brain \emph{without} compromising the meaning and salience of the information to the animal. In animal brains, downsampling is critical because the neural representation of an object or scene must be sparse coded across a neural population if that object if to be effectively learned, discriminated, or recognised.  Sparse coding means that only a very small percentage of the population of neurons engaged in visual learning and recognition in the central brain is activated by any one image.  Without sparse coding, in a finite population of biological neurons the representations of different objects overlap, interfere and cannot be effectively discriminated. This is therefore standard in biological systems. The principle of sparse coding in biology contrasts with common approaches in robotic and computer vision, however, which are typically dense coded to emphasise the preservation of visual information despite accompanying high compute costs. Since many animals, including humans, can rapidly learn to recognise visual stimuli and scenes, there is no obvious performance costs to the sparse coding found in biology.

Fast-flying insects, such as dragonflies, flies and bees rely on vision for flight control, predator avoidance, and for some species prey capture. To respond appropriately to such diverse visual stimuli, their brains need to assess them, so that a rapidly approaching predator is avoided, but a potential food source approached. These tasks are solved by the insect central brain, which is tiny, but highly effective.  A honey bee brain contains less than one million cells, and yet bees are competent and rapid visual learners~\cite{MaBouDi2020, MaBouDi2025}.  They are highly proficient in learning complex multipart, or configural visual stimuli, even human faces~\cite{Dyer2005, Avarguès-Weber2010, Giurfa2012}. Bees are competent visual navigators, able to use learned scenes to pilot a route home~\cite{Freas2023}.  And bees can rapidly learn to distinguish and classify flowers based on their appearance~\cite{HempeldeIbarra2022}. This type of visual learning requires the mushroom body of the insect brain~\cite{Giurfa2012}.  In bees, highly processed visual input projects from the optic lobes to the input of the mushroom body in a sparse code~\cite{Paulk2008}. The mushroom body itself is composed of thousands of Kenyon cells. A visual input results in an extremely sparse activation of Kenyon cells, and this sparse coding across the Kenyon cell population is essential for the mushroom body to operate effectively, and for visual learning to be achieved~\cite{Honegger2011, Ganguly2024}.  

In honey bees the visual input projected to the mushroom body is estimated to be spatially downsampled to approximately $1\%$  of the input to the retina~\cite{MaBouDi2025}, which illustrates the scale of image downsampling. Downsampling is achieved by a sequence of visual processing modules between the retina and the mushroom body. Visual signals from the retina  are first transformed in the lamina, which emphasizes local spatial contrast~\cite{Stockl2020}. They then propagate to the medulla, where chromatic and achromatic components are computed and integrated~\cite{Paulk2009}, before passing to the lobula for further refinement into higher-order feature representations~\cite{Paulk2008-2}. Lobula outputs pass via visual projection neurons to the mushroom body, which provide the primary visual input to the Kenyon cells of the mushroom body~\cite{Wu2016}.

Models of the insect mushroom body have been developed for visual scene learning~\cite{Ardin2016}, colour learning~\cite{MaBouDi2020}, and visual discrimination and classification~\cite{Cope2018}, but they are quite bespoke and operate for the images and task for which they were developed only~\cite{Fu2019, Gkanias2019, MaBouDi2025}. In addition, and in contrast to many models for motion vision~\cite{Fu2019, Fu2020, Ghosh2025}, they often ignore the sequential processing in the insect optic lobes. To redress this, we here provide a model of insect visual processing that is generalisable enough to be used as a "vision module" for other researchers modelling neuromorphic applications, or the insect brain. In order to achieve this, we have chosen to model each module of insect visual processing as a convolutional neural net (CNN) and a custom designed sparse Kenyon cell output. In this paradigm, we have allowed there to be training of the model by backpropagation and gradient descent.  By this choice we have deviated from modelling the biological computations within each module of the insect visual system, and instead use bio-inspired principles with conventional machine learning methods. Whether insect brains could implement backpropagation and generate global error signals as used in neural networks is a topic of ongoing debate~\cite{Lillicrap2020}. It is also unclear to what degree there is experience dependent plasticity within the visual processing modules.  However, by allowing this deviation from a strict attempt to model insect neural circuits we have been able to produce a model of insect visual processing that can be generalised to any image or scene without having to manually adjust the code.  

Our model highlights the effectiveness and efficiency of the insect visual system, and also provides a generalisable model of neuromorphic sparse visual processing that could be applied in neuromorphic vision models, or to models of insect learning or decision making. We have developed both artificial neural network (ANN) and a neuromorphic spiking neural network (SNN) variants of the model for the benefit of the research community.

    \begin{figure}[t]
    \centering
        \includegraphics[width=\textwidth]{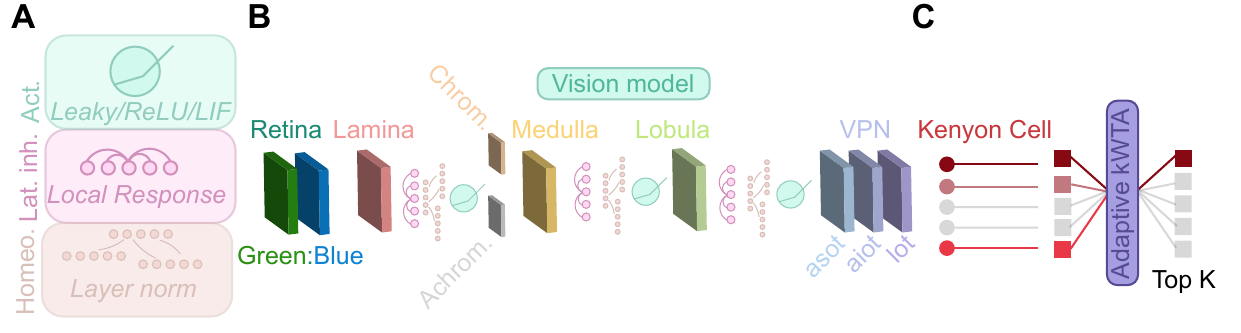}
        \caption{\textbf{Overview of the structure and training of the vision system.} \textbf{A} Activation (Act.) with homeostatic (homeo.) and lateral inhibition (lat. inh.) normalization layers were implemented for \textbf{B} each layer of the CNN, emulating biological processes, including separate chromatic (chrom.) and achromatic (achrom.) processing. \textbf{C} The output is a sparse, 1,024 dimensional linear Kenyon Cell (KC) code from the visual projection neurons (VPN) formed from the anterior superior optic tract( asot), anterior inferior optic tract (aiot), and lateral optic tract (lot) -- representing the \emph{Top K} activations from a given input photoreceptor.}
    \label{fig:vizmodule}
    \end{figure}

\section*{Materials and methods}

\label{subsec:vizmodule}
The insect inspired vision model was developed as a series of convolutional neural network (CNN) layers (Fig.~\ref{fig:vizmodule}). Each of the CNN layers captures the principles of processing performed by the different layers of the insect visual system (Fig.~\ref{fig:vizmodule}B). The vision model $vision$ performs the complete layer-by-layer processing of input images to reduce them to a sparse Kenyon cell $KC$ representation $\mathbb{I}_{in}$:

    \begin{align}
    \begin{gathered}
        KC = vision(\mathbb{I}_{in}).
    \label{eq:vision}
    \end{gathered}
    \end{align}

\subsection*{Convolutional neural network structure and implementation}
Each of the processing layers $\mathcal{PL}$ in $vision$ consists of a 2-dimensional convolutional layer $\mathbb{CNN}$ and an Activation–Normalization block $\ANBlock$ that performs feature extraction and non-linear activation with homeostatic regularization, respectively. A $\mathcal{PL}$ can generally be described by:

    \begin{align}
    \begin{gathered}
        \mathbb{I}_{conv} = \mathbf{CNN}(\mathbb{I}_{in}),\\
        \mathbb{I}_{out} = \ANBlock(\mathbb{I}_{conv}).
    \label{eq:proclayer}
    \end{gathered}
    \end{align}
    where $\mathbb{I}_{conv}$ is the convolved output of an input image and $\mathbb{I}_{out}$ is the processed output of a $\mathcal{PL}$.

For the $\ANBlock$, we employ a single activation function with two normalization strategies to approximate local and global neuronal homeostatic mechanisms inspired by neural circuits~\cite{Carandini2012, Krizhevsky2017}. Our activation function $\mathbb{A}$ is a LeakyReLU function to allow both positive and negative values to flow through the model:

    \begin{align}
    \begin{gathered}
        \mathbb{A}(\mathbb{I}_{conv}) = \max(0,\mathbb{I}_{conv}) + slope \cdot \min(0,\mathbb{I}_{conv}).
    \label{eq:leaky}
    \end{gathered}
    \end{align}
    where $slope$ is a constant that controls the angle of negative slope and produces an activated convolved output $\mathbb{I}_{conv}^{act}$.

To approximate homeostatic neuronal regularization across layers~\cite{Carandini2012}, we utilize two normalizations of outputs. The first is a LocalResponseNorm $\mathbb{N}_{local}$ that functions similarly to the lateral inhibition of spatially neighboring neurons to produce a locally normalized output $\mathbb{I}_{local}$~\cite{Krizhevsky2017}. The purpose of $\mathbb{N}_{local}$ is to regularize neuronal activity and also enhance contrast in the output. It is defined as:

    \begin{align}
    \begin{gathered}
        \mathbb{I}_{local} = \mathbb{I}_{conv}^{act}(k + \frac{\alpha}{n} \sum_{c'=\max(0, c-n/2)}^{\min(N-1,c+n/2)}a_{c'}^2)^{-\beta}
    \label{eq:localnorm}
    \end{gathered}
    \end{align}
    where $k=1$ and is a constant additive factor, $\alpha=0.0001$, $\beta=0.75$, $N$ is the number of input images or batch size, $c$ is the input channel, and $n$ is the number of neighbouring channels to normalize across.

Global normalization across extracted feature channels is provided by GroupNorm $\mathbb{N}_{group}$ and prevents any single feature channel from dominating the processed output. $\mathbb{N}_{group}$ uses normalization strategies that are inspired by processes in biology, performing single observation normalization with local information~\cite{Wu2018}. This is in contrast to commonly used batch normalization strategies in neural networks that normalize over independent samples; a phenomenon not observed in biological systems~\cite{Ioffe2015, Shaw2020, Shen2021}. $\mathbb{N}_{group}$ ensures that diversity in feature extraction is achieved, which is necessary for the generalisability of the $vision$ model. $\mathbb{N}_{group}$ is defined as:

    \begin{align}
    \begin{gathered}
         \mathbb{I}_{out} = \frac{\mathbb{I}_{local} - \mathrm{E}[\mathbb{I}_{local}]}{ \sqrt{\mathrm{Var}[\mathbb{I}_{local}] + \epsilon}} * \gamma + \delta
    \label{eq:groupnorm}
    \end{gathered}
    \end{align}
    where $\mathrm{E}$ is the mean of the current group's channels and spatial dimensions for $\mathbb{I}_{local}$, $\mathrm{Var}$ is the variance, $\gamma$ and $\delta$ are learnable per-channel affine transform parameters, and $\epsilon=0.00001$ is a constant stability factor.

\subsection*{Processing steps in \emph{vision}}
Processing of input images $\mathbb{I}_{in}$ involves multiple $\mathcal{PL}$ steps, that each captures the principles of the different layers of processing in the insect visual system. We describe the implementation of each $\mathcal{PL}$ below. Early layers, such as the retina, process dense visual inputs emulating the high amount of photoreceptor input to visual systems (Fig.~\ref{fig:ann}A, left). Downstream layers, such as the medulla and lobula, refine dense input to create a sparse output by learning and processing important features in the input.

$\mathbb{I}_{in}$ is initially represented by a retinal layer $\mathcal{PL}_{ret}$ that processes the blue:green input images with a $\mathbf{CNN}$. We chose to model just two photoreceptor channels for simplicity and to reduce computational overhead. $\mathcal{PL}_{ret}$ is:

    \begin{align}
    \begin{gathered}
        \mathbb{I}_{ret} = \mathcal{PL}_{ret}(\mathbb{I}_{in}),\\
        \mathbb{I}_{lam} = [\mathbb{I}_{ret} \Vert -\mathbb{I   }_{ret}],
    \label{eq:retina}
    \end{gathered}
    \end{align}
    where $\mathbb{I}_{lam}$ is the input to the lamina and is a concatenation of the output $\mathbb{I}_{ret}$ and a negative transform $-\mathbb{I}_{ret}$ from $\mathcal{PL}_{ret}$, to approximate the convergence of biological positive- and negative-contrast pathways in the compound eyes of insects~\cite{Soto2025}. 

$\mathbb{I}_{lam}$ is subsequently processed by $\mathcal{PL}_{lam}$, which approximates the spatial contrast enhancement in the lamina through the training we allowed in the $\mathcal{PL}_{lam}$ layer. We achieve the spatial contrast enhancement through the positive and negative representation of the $\mathbb{I}_{ret}$, rather than defining a specific convolutional process to achieve this:

    \begin{align}
    \begin{gathered}
        \mathbb{I}_{med} = \mathcal{PL}_{lam}(\mathbb{I}_{lam}),\\
    \label{eq:lamina}
    \end{gathered}
    \end{align}

$\mathcal{PL}_{lam}$ generates the lamina output $\mathbb{I}_{med}$ which is then processed by the medulla layer $\mathcal{PL}_{med}$. This processes both chromatic and achromatic features separately and in parallel  by enhancing the most prominent spatial elements from a color and a greyscale representation of the $\mathbb{I}_{lam}$, respectively. In our model that enhancement was achieved by the training we allowed within the $\mathcal{PL}_{med}$ layer.
    
    \begin{align}
    \begin{gathered}
        \mathbb{I}_{lob} = \mathcal{PL}_{med}(\mathbb{I}_{med}^{c} \Vert \mathbb{I}_{med}^{a}), 
    \label{eq:medulla}
    \end{gathered}
    \end{align}
    where $\mathbb{I}_{lam}^{a}$ is the mean across the blue and green color channels of $\mathbb{I}_{lam}^{c}$ to generate a greyscale representation such that achromatic features can be learned. After processing through the medulla, chromatic and achromatic features are  processed by the lobula layer $\mathcal{PR}_{lob}$ which is a further processing layer to refine the medulla output: 

    \begin{align}
    \begin{gathered}
        \mathbb{I}_{vpn} = \mathcal{PR}_{lob}(\mathbb{I}_{lob}),
    \label{eq:lobula}
    \end{gathered}
    \end{align}

We then split $\mathbb{I}_{vpn}$ into three distinct projection neuron $\mathbb{VPN}$ pathways---capturing the insects anterior superior optic tract $\mathbb{I}_{asot}$, anterior inferior optic tract $\mathbb{I}_{aiot}$, and lateral optic tract $\mathbb{I}_{lot}$ to create an input $\mathbb{VPN}$ to be processed into a sparse Kenyon cell representation. Each of the three distinct $\mathbb{VPN}$ pathways processes different elements and spatial features captured by the lobula layer, to enhance and improve unique and diverse representations for KC processing. 

    \begin{align}
    \begin{gathered}
        \mathbb{VPN} = \mathbb{I}_{asot} \Vert \mathbb{I}_{aiot} \Vert \mathbb{I}_{lot},
    \label{eq:vpn}
    \end{gathered}
    \end{align}

Before passing the $\mathbb{VPN}$ outputs to the KC layer for sparse activation, we needed to reduce the number of channel dimensions from the CNN by performing an average pooling function. In CNNs, features from images are detected and split into individual channels during the processing. When more channels are used, more unique features can be discovered during the output of each layer. However, too many feature channels can increase the computational load to levels beyond what conventional compute platforms are able to handle. In our case, the total number of feature channels from our CNN is 128. The average pooling process lowers this to a single combined feature channel that eases computational load and generates a single representation that can be used for the $KC$ population. 

\subsection*{Sparse Kenyon cell activations}
\label{subsec:sprasekc}
In this section, we describe the generation of the sparse $KC$ representation that is the output of the $vision$ model and is assessed in our downstream evaluations. Since CNNs are typically considered a dense computation, we developed a mechanism to turn dense linear outputs from a CNN into a sparse ($\approx$5\% active) representation through output masking, activity thresholding, and activity adjustment. The final $KC$ output layer produces a very sparse pattern of activation from the output of the lobula layer and provides a low dimensionality representation of higher order inputs (Fig.~\ref{fig:ann}B).

We created a custom activation linear function $Sparse$ to approximate the sparsity of the representation $KC$~\cite{Ganguly2024}:

    \begin{align}
    \begin{gathered}
        KC = Sparse(\mathbb{VPN}),
    \label{eq:sprasevpn}
    \end{gathered}
    \end{align}
    where the $Sparse$ transformation is defined as:

    \begin{align}
    \begin{gathered}
        KC = \mathbb{VPN}(\mathsf{W} \odot \mathsf{M})^{\intercal} + b,
    \label{eq:sparsefunc}
    \end{gathered}
    \end{align}
    where $\mathsf{W}$ is the tunable weight matrix, $\mathsf{M}$ is a fixed binary connectivity mask between ${0,1}$, $\intercal$ is a matrix transpose, $b$ is an optional bias vector, and $\odot$ is the Hadamard product.

The connectivity mask $\mathsf{M}$ for neuron $i$ to neuron $j$ is determined from a random sample $S_i$:

    \begin{align}
    \begin{gathered}
        \mathsf{M}_{ij} = \begin{cases}
            1 & \text{if } j \in S_i \\
            0 & \text{otherwise},
        \end{cases}
    \label{eq:binmatrix}
    \end{gathered}
    \end{align}

$\mathsf{M}$ simply defines a connectivity pattern that creates sparse connections in an otherwise dense neural network linear layer. Therefore, we additionally applied an adaptive k-Winner-Take-All (a-kWTA) function to competitively discover the highest responding outputs from the $KC$ to further sparsify outputs, capturing meaningful outputs from complex visual stimuli with as few as 30 neurons.

The a-kWTA function also assists in regularizing the activity of the $KC$ representation, encouraging diversity in feature outputs and preventing single neurons from dominating. We utilize this function for both training efficacy and downstream evaluation. a-kWTA runs three sequential components: 1) activity frequency tracking, 2) adaptive threshold \& activation adjustment, and 3) k-WTA selection.

For $KC$ output, activity frequency tracking calculates the running mean neuron activation frequencies $\mu$. This is tracked using an exponential moving average in order to update the thresholds for the activation of a neuron:

    \begin{align}
    \begin{gathered}
        \mu_t^{(i)} = \alpha \mu_{t-1}^{(i)} + (1-\alpha) \cdot \frac{1}{B} \sum_{b=1}^{B} \mathbb{I}(KC_b^{(i)} > 0),
    \label{eq:runningmean}
    \end{gathered}
    \end{align}
    where $\mu_t^{i}$ is the running activation frequency for neuron $i$ at time $t$, $\alpha$ is a momentum parameter specifying overactivity to be observed across several timesteps, $B$ is the batch size, and $\mathbb{I}(KC_b^{(i)} > 0)$ is an indicator function that binarizes whether or not a neuron had any activity to only track active neurons.

The threshold $\theta$ for individual neurons is adaptively modulated by the updated $\mu$ running mean from Eq. \ref{eq:runningmean} and its deviation from a desired target sparsity value $\rho$. $\theta$ is:

    \begin{align}
    \begin{gathered}
        \theta^{(i)} = 1 + 2 \cdot \max(0, \mu_t^{(i)} - \rho),
    \label{eq:threshold}
    \end{gathered}
    \end{align}
    where the value of $\theta$ is clamped to a $\min = 0$. The activity of $KC$ is then modulated and regularized to produce a $KC_{adj}$ by the updated $\theta$ values from Eq. \ref{eq:threshold}: 
    
    \begin{align}
    \begin{gathered}
        KC_{adj} = \frac{KC}{\theta},
    \label{eq:adjustment}
    \end{gathered}
    \end{align}

Once $KC_{adj}$ has been generated, we then select the indexes of the top neurons $idx$ based on their adjusted activations:
    
    \begin{align}
    \begin{gathered}
        idx = \text{TopK}(KC_{adj}, \max(1, \lfloor \rho \cdot {dim_{KC}} \rfloor)),
    \label{eq:topkselection}
    \end{gathered}
    \end{align}
    where $dim_{KC}$ is the dimensionality of the $KC$ projection and $k$ is the determinate of the number of top K results to return.

Lastly, the final sparse and regularized $KC$ is returned as:

    \begin{align}
    \begin{gathered}
        KC^{(i)} = \begin{cases}
        KC^{(i)} & \text{if } i \in idx \\
        0 & \text{otherwise}.
    \end{cases}
    \label{eq:finalkc}
    \end{gathered}
    \end{align}

\subsection*{Spiking neural network implementation}
\label{subsec:snn}
In computational models of biological systems, SNNs are a neuromorphic counterpart to the ANN, such as the one presented in Fig.~\ref{fig:snn}. To allow for compatibility with multiple downstream modelling systems, we present an SNN version of the $vision$ system.

To implement the spiking neural network (SNN) vision model---$vision^{SNN}$---we replace all instances of the $leaky$ activation layers in the CNN with a leaky integrate-and-fire (LIF) neuron model:

    \begin{align}
    \begin{gathered}
        \tau \frac{dU_{mem}(t)}{dt} = -U_{mem}(t) + RI_{in}(t).
    \label{eq:lif}
    \end{gathered}
    \end{align}
    where $\tau$ is the time constant, $U_{mem}$ is the membrane potential, $R$ is the resistance, $I$ is the current, and $t$ is the timestep. 

Since the SNN iterates inputs over time, we generate Bernoulli spikes $B_{spk}$ of input images $\mathbb{I}_{in}$ for each $t$:

    \begin{align}
    \begin{gathered}
        B_{spk}(t) = \mathbf{1}_{\{u < \mathbb{I}_{in}(t) \cdot \rho\}}.
    \label{eq:bernoulli}
    \end{gathered}
    \end{align}
    where $u$ is a random variable drawn from a uniform distribution between 0 and 1 and $\rho$ is a spike rate scale constant.

In $vision$, we required the use of a k-WTA function to approximate the sparsity of the $KC$. For $vision^{SNN}$, sparse and spiking $KC^{SNN}$ activations are learnt by tuning the dynamics and activation thresholds of spiking LIF neurons, rather than enforcing sparsity through k-WTA as in $vision$. Therefore, we replace the $k-WTA$ calculation for $KC$ with a single $LIF$ layer after producing the $\mathbb{T}_{VPN}$:

    \begin{align}
    \begin{gathered}
        KC^{SNN} = LIF(\mathbb{T}_{VPN}).
    \label{eq:kcsnn}
    \end{gathered}
    \end{align}
    
\subsection*{Model training}
\label{subsec:modeltrain}
\begin{figure}[t]
    \centering
        \includegraphics[width=\textwidth]{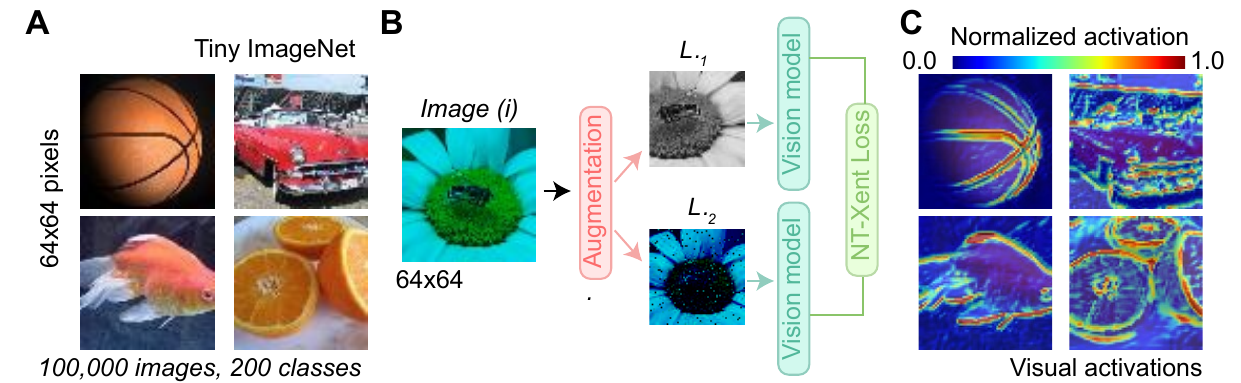}
        \caption{\textbf{Visual model training using self-supervised contrastive learning.} \textbf{A} Example images from the Tiny ImageNet dataset~\cite{Le2015}, which consists of 100,000 64x64 pixel images from 200 different data classes. \textbf{B} Schematic example of the SimCLR training method~\cite{Chen2020}. Images from the Tiny ImageNet dataset are converted to blue:green images and have two separately random augmentations applied, which are then subsequently processed through the visual model (Fig.~\ref{fig:vizmodule}), where finally an NT-Xent loss function is calculated to minimize the distance between the KC representations that $vision$ produces. \textbf{C} Example activations from a fully trained model, showing which visual features $vision$ responds strongest to.}
    \label{fig:training}
    \end{figure}

So that $vision$ and $vision^{SNN}$ could be applied to any image set we used training of the CNNs to tune the model weights to respond to salient information. In contrast to training neural networks in the traditional sense, where often the goal is to classify or identify specific features, we use a self-supervised paradigm that allows the model to learn to produce distinct $KC$ representations resulting from a wide variety of visual activations in distinct images (Fig.~\ref{fig:training}).

To tune both $vision$ and and $vision^{SNN}$  we use the normalized temperature-scaled cross-entropy (NT-Xent) loss function $\mathcal{L}_{NT-Xent}$ which provides contrastive learning~\cite{Sohn2016, Wu2018-2, Oord2019, Chen2020}, as implemented in SimCLR~\cite{Chen2020}. In SimCLR training, two augmentations $\mathcal{A}$ from an original input are generated and processed through $Sparse$ with the goal to maximize the agreement across the normalized output from these functions (Fig.~\ref{fig:vizmodule}B): 

    \begin{align}
    \begin{gathered}
        KC_1^{\mathbb{N}}, KC_2^{\mathbb{N}} = vision(\mathcal{A}^1\mathbb{T}_{in})^{\mathbb{N}}, \text{ } vision(\mathcal{A}^2\mathbb{T}_{in})^{\mathbb{N}},
    \label{eq:kcprojects}
    \end{gathered}
    \end{align}
    where ${\mathbb{N}}$ represents $L_p$ normalization.

These two outputs are then concatenated and used to compute cosine-scaled similarities $S$:

    \begin{align}
    \begin{gathered}
        \mathbf{S} = \frac{KC_1^{\mathbb{N}} \Vert KC_2^{\mathbb{N}} \cdot (KC_1^{\mathbb{N}} \Vert KC_2^{\mathbb{N}})^{\intercal}}{T}, 
    \label{eq:cosinekc}
    \end{gathered}
    \end{align}
    where $(KC_1^{\mathbb{N}} \Vert KC_2^{\mathbb{N}})^{\intercal}$ represents the transposed matrix and $T$ is the temperature variable.

In order to prevent trivial solutions being converged upon during training, self-similarities for row $m$ in column $n$ for $S$ are masked such that they do not contribute to $\mathcal{L}_{NT-Xent}$:

    \begin{align}
    \begin{gathered}
        \tilde{\mathbf{S}}{m,n} = \begin{cases}
        \mathbf{S}{m,n} & \text{if } m \neq n \\
        -\infty & \text{if } m = n
        \end{cases}
    \label{eq:masking}
    \end{gathered}
    \end{align}

Once masked, we identify positive pairs $pos$ in $\mathbf{S}{m,n}$ by:

    \begin{align}
    \begin{gathered}
    pos(i) = \begin{cases}
        i + B & \text{if } i \in {0, 1, ..., B-1} \\
        i - B & \text{if } i \in {B, B+1, ..., 2B-1}
    \end{cases}
    \label{eq:positive_pairs}
    \end{gathered}
    \end{align}
    which effectively pairs each sample $KC$ with its corresponding augmented counterpart in the batch $B$.

\subsection*{Implementation details}
\label{subsec:impdetails}
The $vision$ and $vision^{SNN}$ models are implemented in Python3, PyTorch, and SNNTorch employing a mixture of built-in and custom functions to train and evaluate our models~\cite{Paszke2019, Eshragian2023}. For reproducibility, we have implemented this work using the Pixi environment and package dependency management system, allowing for cross-platform compatibility and bit-for-bit results consistency~\cite{Fischer2025}. $vision$ and $vision^{SNN}$ were trained using a A100 graphics processor unit (GPU) using a High Performance Computing (HPC) cluster. In $vision$ models connection weight tuning was trained using the Tiny-ImageNet dataset~\cite{Le2015} for 20 epochs, with 100,000 training examples, a batch size of 128, and a learning rate of $1 \times 10^{-4}$. $vision^{SNN}$ was trained for 20 epochs, 100,000 training examples, a batch size of 16 with 25 timesteps due to the additional memory requirement, and a learning rate of $1 \times 10^{-4}$. Model parameters were updated using the AdamW optimizer~\cite{Loshchilov2019} using a weight decay of $1 \times 10^{-4}$, with the learning rate set to a cosine annealing schedule that steps at the end of each epoch.

\subsection*{Data and code availability}
\label{subsec:codedata}
All datasets used in this work is freely and publicly available~\cite{Nilsback06, Le2015}. The code and pre-trained model weights for this work are open-sourced and publicly available at: \texttt{https://github.com/AdamDHines/ApiaViz}.

\subsection*{Experiments}
\label{subsec:experiments}
In \textbf{Experiment 1} we tested $vision$ and $vision^{SNN}$ on a simple binary classification task. Five images of lavenders and five images of sunflowers were used (Adobe Creative Commons Photos) and run through both $vision$ and $vision^{SNN}$ to retrieve $KC$ outputs, which were averaged and compared using the cosine similarity function $K$ (Figs.~\ref{fig:ann}-~\ref{fig:snn}): 

    \begin{align}
    \begin{gathered}
        K(X, Y) = \frac{\langle X, Y \rangle}{\lVert X \rVert \, \lVert Y \rVert}
    \label{eq:cosine}
    \end{gathered}
    \end{align}
    where $X/Y$ are two separate $KC$ output matrices concatenated from multiple image presentations. Higher cosine similarity values ($>0.5$) indicate that the $vision$ and $vision^{SNN}$ outputs are more similar to each other, whereas lower cosine similarity ($<0.5$) suggest that the outputs are less similar.

\textbf{Experiment 2} further tested the capabilities of $vision$ and $vision^{SNN}$ to classify flower images by type in the 17 Category Flower Dataset (17CFD), which consists of 17 different flower species with 80 images each across a variety of viewpoints, image quality, and distractors such as animals, insects, and other natural scenery~\cite{Nilsback06}. The key aim of using the 17CFD was to evaluate how well $vision$ and $vision^{SNN}$ could classify natural images.

For $vision$, we developed a temporal scanning paradigm that accumulated $KC$ activity over multiple time windows---simulating an insect moving around the flower. Smaller patches of an image were generated from a randomly determined scanning path (Fig.~\ref{fig:natscene}A) and the $KC$ output from $vision$ were accumulated to produce a running average of activity $\bar{KC}$ over time:

    \begin{align}
    \begin{gathered}
        \bar{KC}_{t+1} = (\bar{KC}_t \cdot \kappa) + KC_t 
    \end{gathered}
    \end{align}
    where $t$ is the timestep and $\kappa$ is a small decay factor set to $\kappa = 0.9$.

All 80 images from each flower type was used with 10 unique scanning patterns per image extracted for linear classification. We took all the $\bar{KC}$ outputs for each image and scanning patterns and trained a small linear classifier---implemented in scikit-learn's LogisticRegression function~\cite{Pedregosa2011}---to evaluate if $vision$ would be able to distinguish the different flower types (Fig.~\ref{fig:natscene}). We employed this testcase to additionally identify training efficiency, in order to understand how quickly $vision$ learns salient information for effective sparse $KC$ representation of images. 

When evaluating $vision^{SNN}$, the spiking output of the $KC$ using the same temporal scanning paradigm for $vision$ was not effective at linear classification. This was due to how information is propagated through spiking networks and single-view patch presentations were insufficient at producing a robust $\bar{KC}$ signal~\cite{Eshragian2023}. Therefore, instead of focusing on classification accuracy we instead assessed how unique the $vision^{SNN}$ output for each flower type was by measuring the $KC$ neuron selectivity. The selectivity index $SI$ was chosen to evaluate the $vision^{SNN}$ output and was calculated as the $90^{th}$ percentile of the neuron-wise selectivity values, which for neuron $i$ is:

    \begin{align}
    \begin{gathered}
        SI_i = \frac{r_{i,max} - \bar{r}_{i,others}}{r_{i,max} - \bar{r}_{i,others} + \epsilon}
    \label{eq:selectivity}
    \end{gathered}
    \end{align}
    where $r_{i,max}$ are the mean neuron responses for the flower class that elicits the highest average firing of that neuron $i$, $\bar{r}_{i,others}$ is the mean firing response of all other neurons, and $\epsilon$ is a small additive factor to prevent division by $0$.

\textbf{Experiment 3} was a Visual Place Recognition (VPR)~\cite{Lowry2016, Schubert2024} task (Fig.~\ref{fig:vpr}). VPR is an image retrieval problem, where a known reference database of images is compared against incoming query images with the highest similarity score considered the recognised place.

We used the Gardens Point Walking dataset, which contains multiple traverses of the $\approx500$ m QUT Gardens Point campus (Fig.~\ref{fig:vpr}A) with a lateral pose shift causing similar salient features to have a different field of view (Fig.~\ref{fig:vpr}B). Reference images were taken from the "Left" traverse and queries from the "Right" traverse and resized to $75\times75$ pixels from before running the vision modules (Fig.~\ref{fig:vpr}B). 

Each reference and query image is one matching pair, as shown in the ground truth matrix (Fig.~\ref{fig:vpr}C). Commonly, a tolerance for VPR is set to account for matches within a set distance of the actual location to alleviate overly strict matching criteria. In our case, we set a ground truth tolerance of $\pm$ 3 places which equate to $\approx15$ m. If a match occurred within this tolerance of the actual reference place, we consider that match to be correct. Across the SAD method and both $vision$ models, the similarity of reference to query images was measured through cosine similarity and produced similarity matrices that aligned closely with the ground truth (Fig.~\ref{fig:vpr}C). 

We measured the Recall@K to assess localization performance of cosine similarity generated features from reference and query images~\cite{Schubert2024}. Recall is defined as the number of true positive $TP$ matches over the number of ground truth positive $GTP$ matches total in a reference-query set:

    \begin{align}
    \begin{gathered}
        Recall = \frac{TP}{GTP}
    \label{eq:recall}
    \end{gathered}
    \end{align}
    Recall@K was measured up to a maximum of $K=25$, where $K=1$ only allows a single reference match per query, and $K=5$ allows the top 5 reference matches per query to be considered as a $TP$~\cite{Schubert2024}.

We compared $vision$ and $vision^{SNN}$ with one of the simplest VPR methodologies, the sum-of-absolute-differences (SAD) which computes pixel-wise differences between reference and query images, with the lowest difference generating the highest similarity score. SAD is calculated using the differences $D$ of the absolute pixel values of a query image to every reference image, where the minimum value is the closest match~\cite{Milford2012}:

    \begin{align}
    \begin{gathered}
        D = \frac{1}{R_xR_y} \sum^{R_x}_{x=0} \sum^{R_y}_{y=0} \vert p_{x,y} - p_{x,y} \vert
    \label{eq:sad}
    \end{gathered}
    \end{align}
    where $x,y$ are the spatial coordinates of the image, $R_x,R_y$ are the dimensions of the downsampled image, and $p$ are the pixel intensity values. The sum-of-absolute-differences (SAD) was chosen as a baseline for its model free localization capability.

\section*{Results}
\label{subsec:overview}
\subsection*{Kenyon cell responses to flower images}

    \begin{figure}[t]
    \centering
        \includegraphics[width=\textwidth]{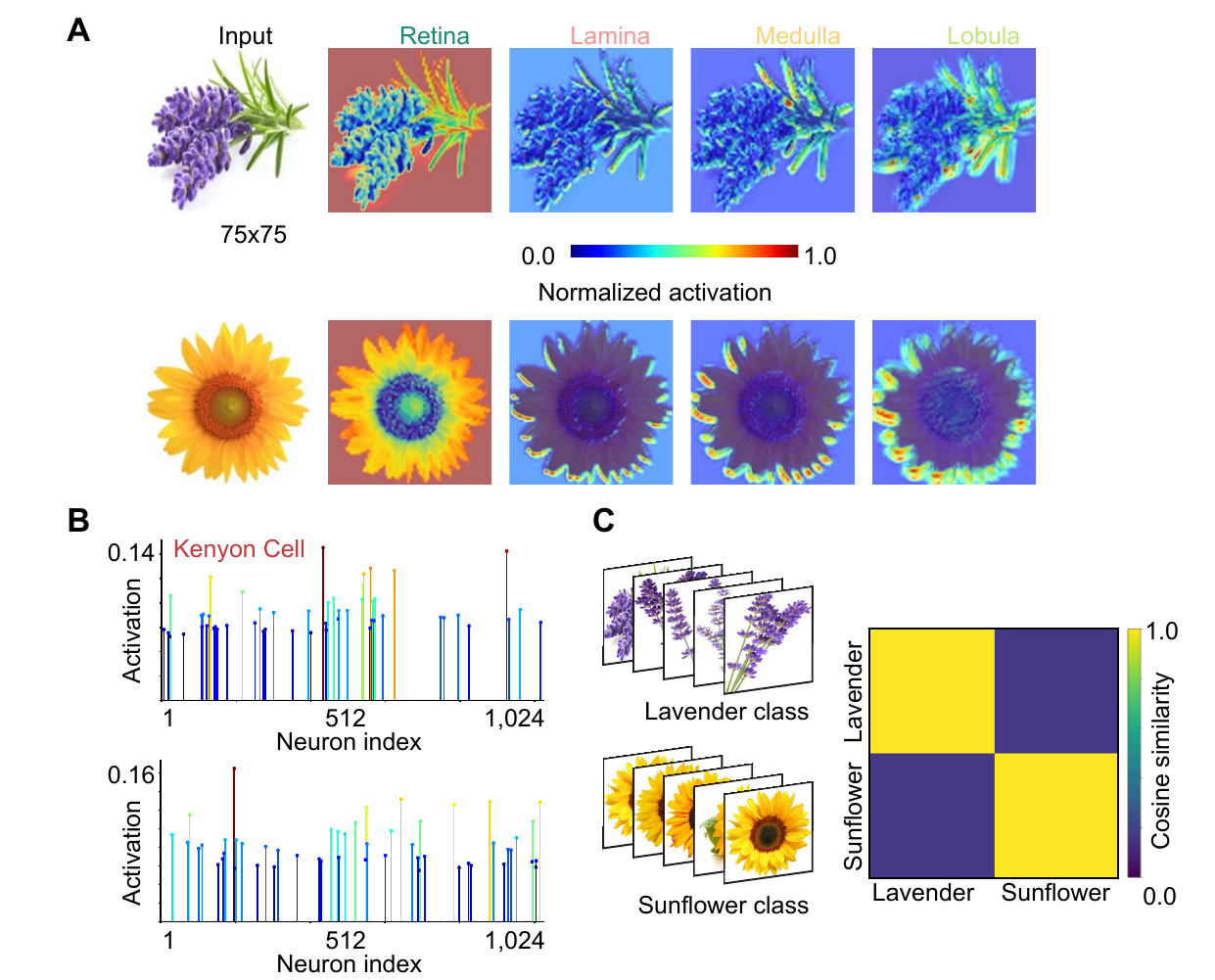}
        \caption{\textbf{The ANN version of \emph produces low-dimensionality Kenyon cell representations.} \textbf{A} Example activations of each layer of the vision model for two different flower inputs (lavender and sunflower), highlighting activations from the CNN. \textbf{B} Kenyon cell representations from the lavender (top) and sunflower (bottom) produces distinct activation patterns that allow them to be \textbf{C} separated and distinguished from one another reliably. The $KC$ has 1,024 output neurons, with each neuron index representing a single $KC$ neuron.}
    \label{fig:ann}
    \end{figure}

\textbf{Experiment 1} compared $KC$ responses to two small datasets of lavenders and sunflowers with 5 example images each (Fig.~\ref{fig:ann}). Fig.~\ref{fig:ann}A-B visually shows the response from each layer of $vision$ to the input flower image and the sparse $KC$ output of $\approx50$ neurons. $vision$ identified common features within flower classes, with these features showing through a low cosine similarity across flower types---$18\%$ (Fig.~\ref{fig:ann}C). 

    \begin{figure}
    \centering
        \includegraphics[width=\textwidth]{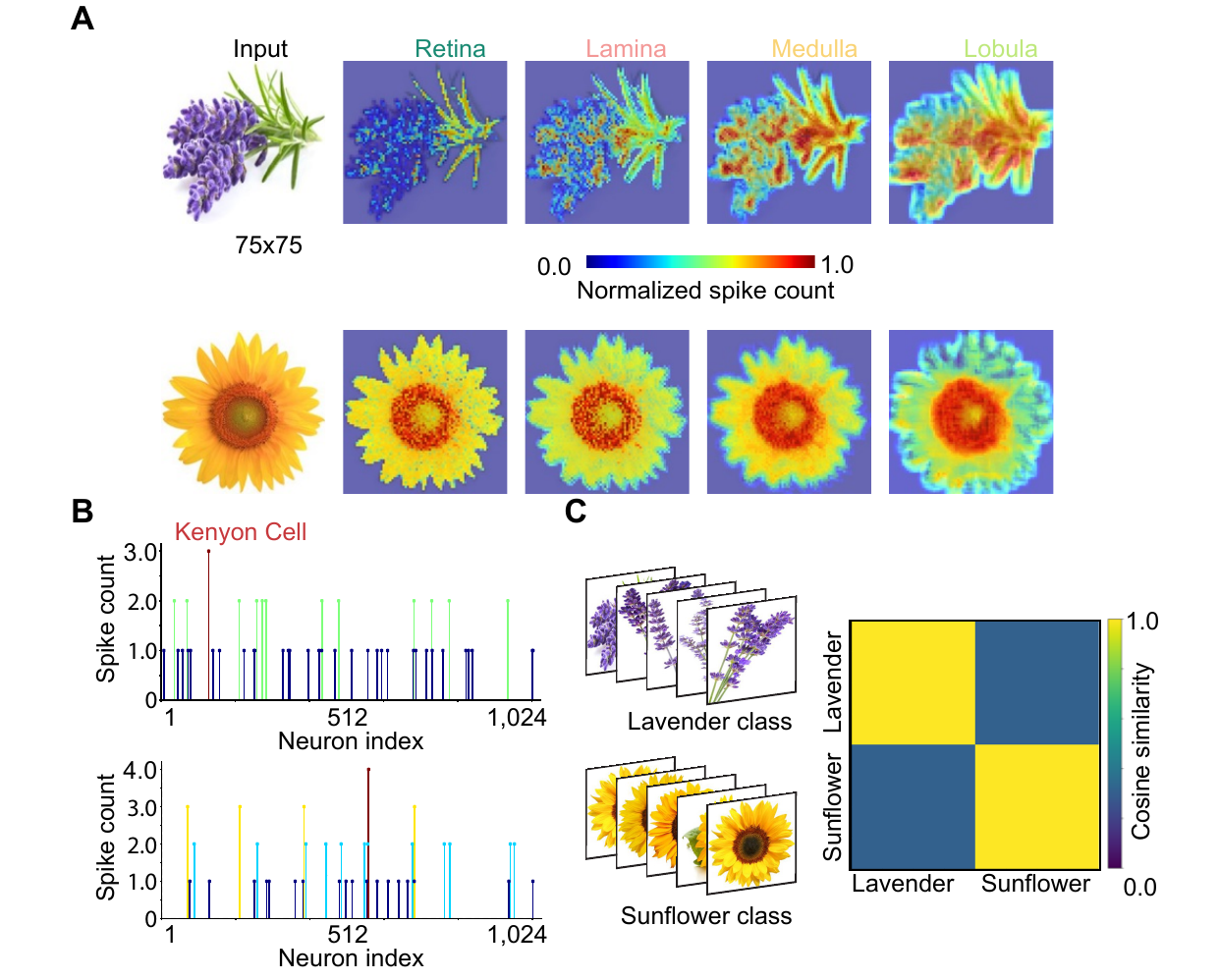}
        \caption{\textbf{The SNN version of vision creates sparse Kenyon cell codes through spiking neuron dynamics.} \textbf{A} Representative spiking activations from each layer of the SNN for a 75x75 pixel input of a lavender and sunflower, \textbf{B} showing varied sparse KC representations. \textbf{C} The SNN is capable of creating unique KC codes for individual flower classes measured through cosine similarity of KC features.}
    \label{fig:snn}
    \end{figure}

$vision^{SNN}$ was also tested on the same images with the layer activation patterns shown in (Fig.~\ref{fig:snn}A). $KC$ output was also sparse---$\approx50$ neurons active total---and distinct across the flower groups (Fig.~\ref{fig:snn}B). Discrimination of lavenders and sunflowers remained robust in $vision^{SNN}$, with low cosine similarity between flower types---$36\%$ (Fig.~\ref{fig:snn}C). The class separation in $vision$ (Fig.~\ref{fig:ann}C) is better than $vision^{SNN}$, which can be attributed to the commonly observed drop in performance for spiking versions of artificial networks~\cite{Eshragian2023}.

\begin{figure}
    \centering
        \includegraphics[width=\textwidth]{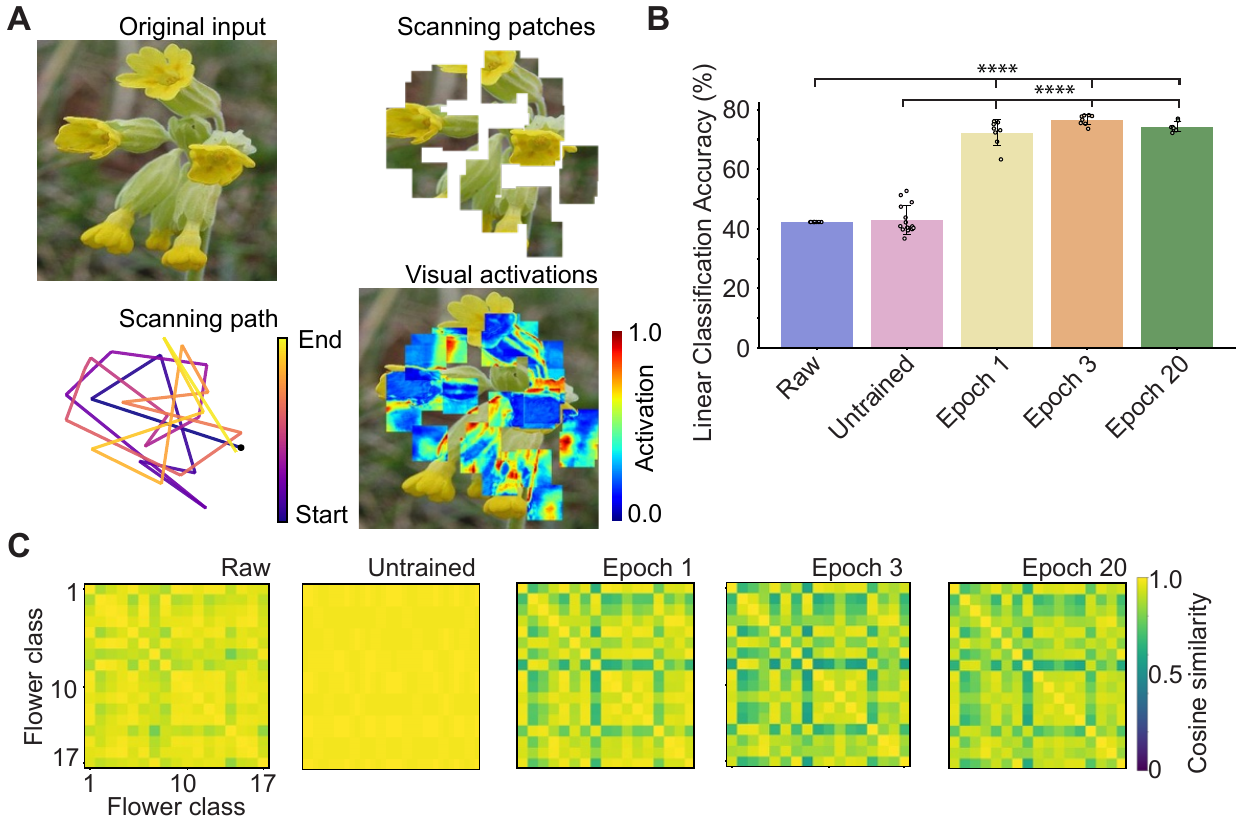}
        \caption{\textbf{Temporal accumulation of information allows for complex image classification.} \textbf{A} Example input image from the 17 Category Flower dataset~\cite{Nilsback06} in its original resolution, with a scanning path of 75x75 pixel patches used to generate temporally accumulated information. \textbf{B} Comparison of linear classification performance of an artificial network in predicting the correct flower species from scanning activations. Raw image inputs to a linear classifier, without any visual processing, performed as well as the untrained vision model (n=8, p=0.989, One-way ANOVA with Tukey's HSD). Training over a single epoch produces accuracy over 70\%, a 30\% increase over untrained or raw inputs, with the peak of 76.6\% after 3 training epochs (n=8, p$<$0.0001, One-way ANOVA with Tukey's HSD). \textbf{C} Cosine similarity matrices of raw pixel similarity and KC output for 17 species of flowers used in \textbf{B}.} 
    \label{fig:natscene}
    \end{figure}

\textbf{Experiment 2} used the 17CFD dataset. $vision$ was compared to raw low-bandwidth image inputs and naive models and correctly classified the different flower types with an accuracy of $40\%$ to $>75\%$ in a single epoch of learning (Fig.~\ref{fig:natscene}B). Raw image inputs and untrained models show an extremely high degree of interclass similarity across all flowers, failing to disambiguate any meaningful features, whereas in 3 training epochs $vision$ learned to disambiguate distinct features from all flower classes (Fig.~\ref{fig:natscene}C). Interestingly, training beyond a single epoch does not yield further improvement in classification performance (Fig.~\ref{fig:natscene}B).
    
    \begin{figure}
    \centering
        \includegraphics[width=\textwidth]{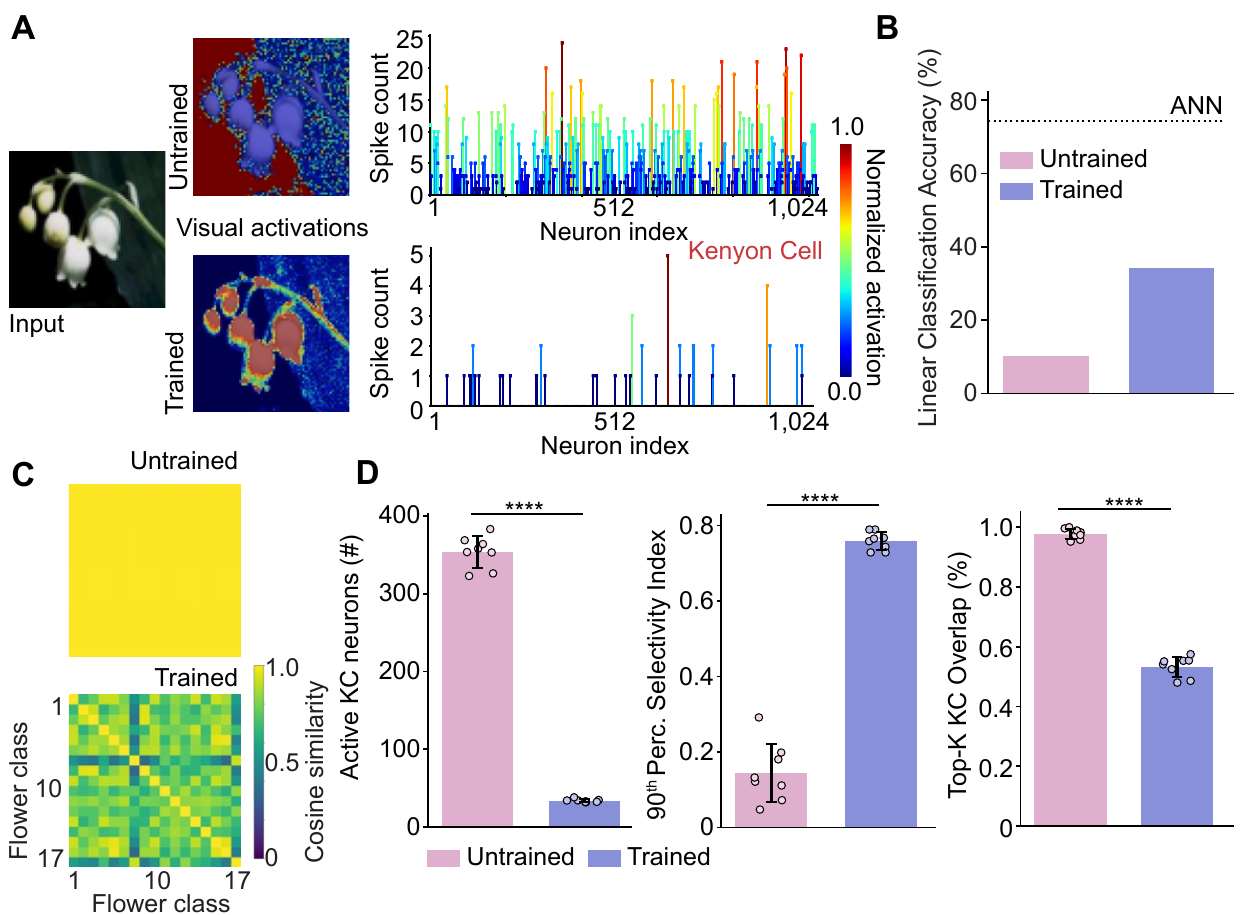}
        \caption{\textbf{The spiking vision model produces selective, sparse activity in response to different flowers.} \textbf{A} An example input image from the 17 Category Flower Dataset (17CFD) resized to 75x75 pixels and the retinal response in an untrained and trained neuromorphic vision model producing dense and sparse $KC$ activation, respectively. \textbf{B} Trained spiking models where unable to achieve linear classification performance relative to the artificial system (ANN). \textbf{C} Cosine similarity for the 17 different flower classes showed strong intraclass similarity and lower interclass similarity. \textbf{D} Analysis of the spiking $KC$ output between untrained and trained models showed a significantly reduced active population from an average of 353 to 33 active neurons when presented with images from the 17CFD (n=8, p$<$0.0001, One-way ANOVA with Tukey's HSD). Although output spikes were not linearly classifiable, we observed a significant increase in the selectivity index of each output Kenyon cell neuron to specific flower data classes from 14\% to 76\% (n=8, p$<$0.0001, One-way ANOVA with Tukey's HSD). Additionally, the overlap in the Top-K active neurons for each class was significantly reduced from 98\% to 53\% (n=8, p$<$0.0001, One-way ANOVA with Tukey's HSD).}
    \label{fig:snnselectivity}
    \end{figure}

Experiment 2 also tested $vision^{SNN}$ with the 17CFD (Fig.~\ref{fig:snnselectivity}). When comparing naive to trained models using our contrastive learning paradigm, the most noticeable difference was in the $KC$ neuron response which was dense for naive and sparse for trained (Fig.~\ref{fig:snnselectivity}A). Although $vision^{SNN}$ learned sparse representations, the spiking output was not able to be effectively linearly classified when presented with the 17CFD in either scanning presentations (Fig.~\ref{fig:snnselectivity}B). The highest accuracy $vision^{SNN}$ was able to achieve was $\approx 40\%$ using static images, relative to $vision$ that achieved $75\%$ accuracy (Fig.~\ref{fig:snnselectivity}B). The accuracy increase from naive, however, was approximately the same as $vision$ which was $\approx30\%$ (Fig.~\ref{fig:snnselectivity}B).

The cosine similarity matrix, however, showed that when $vision^{SNN}$ was trained it showed strong intra-flower-type similarity and lower inter-flower-type similarity, in comparison to the naive condition (Fig.~\ref{fig:snnselectivity}C). Therefore, we analyzed the spiking activity itself to observe if there was any sparse selectivity in the Kenyon cell responses and neuron activations to each flower species (Fig.~\ref{fig:snnselectivity}D). We observed a significant decrease in the density of active Kenyon cell neurons on average, from 353 to 33 neurons (Fig.~\ref{fig:snnselectivity}D). We observed a significant increase in the selectivity index of these neurons to distinguish individual flower species from $14\%$ to $76\%$  Fig.~\ref{fig:snnselectivity}D). Finally, we also observed that of the Top-K (where $K = 32$) Kenyon cell neurons there was a reduction in the overlap of active neurons across the different flower classes from $98\%$ to $55\%$, indicating that the trained spiking model shows sparse and selective responses to unique flowers.

\label{subsec:navvpr}

    \begin{figure}
    \centering
        \includegraphics[width=\textwidth]{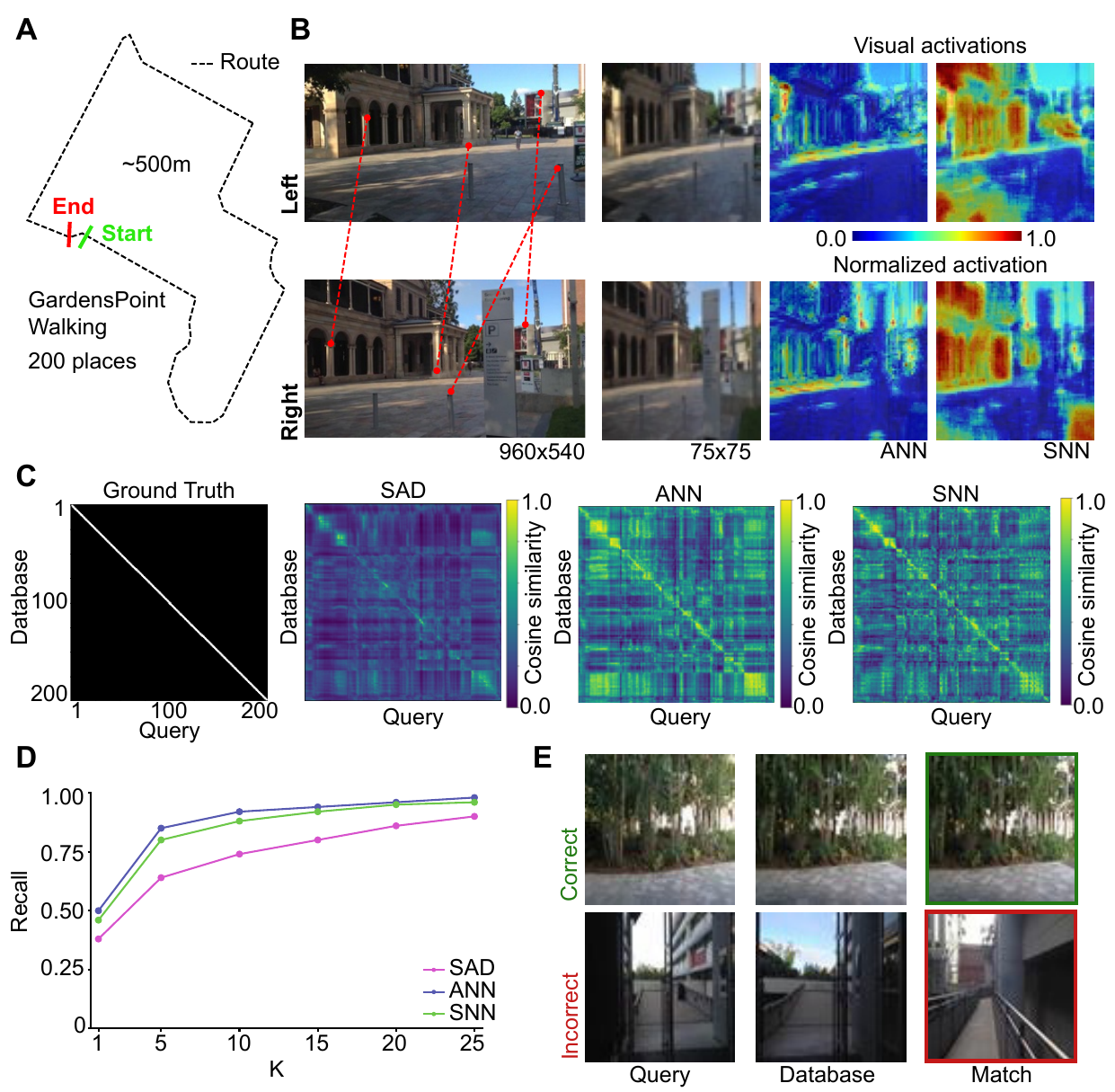}
        \caption{\textbf{Vision model performing a simple localization task in an outdoor environment.} \textbf{A} Schematic of the route from the Gardens Point Walking dataset, which consists of two $\approx500m$ traverses. \textbf{B} An example of a database and query image, in its original 960x540 pixel format. The red lines indicate the same landmark features, distributed across the two sets of images. The 75x75 pixel input is used for processing, where "Left" and "Right" and the two traverses of the same route with a lateral pose shift. Normalized visual activations are highlighted to showcase similar areas of the scene are detected despite different lateral poses. \textbf{C} The ground truth and similarity matrices across the database and query images. The ground truth shows the true correspondence between the query and database images. Similarity matrices calculated through the cosine function indicate how similar a query image is to all reference database images, with higher similarity indicating a stronger match. \textbf{D} Compared to SAD, both the ANN and SNN model showed greater localization accuracy across all Recall@K metrics, where we have evaluated K to the top 1, 5, 10, 15, 20, and 25 places. \textbf{E} Shows an example of a correctly and incorrectly identified place from the ANN.}
    \label{fig:vpr}
    \end{figure}

\textbf{Experiment 3} tested $vision$ and $vision^{SNN}$ on a VPR task (Fig.~\ref{fig:vpr}). Activation heatmaps of identified features from the lobula layer of both $vision$ and $vision^{SNN}$ showed that it is appropriately attending to similar areas of an input image despite lateral pose shifts (Fig.~\ref{fig:vpr}B). Visual features for both $vision$ and $vision^{SNN}$ produced cosine similarity matrices that followed the ground truth closely (Fig.~\ref{fig:vpr}C), with a higher similarity score between corresponding matches in the query and reference database.
 
When compared to the SAD method, $vision$ and $vision^{SNN}$ provided a more accurate estimate of location across the Recall@K metrics (Fig.~\ref{fig:vpr}D). We observed $10\%$ increase in recall using $vision$ and $vision^{SNN}$ (Fig.~\ref{fig:vpr}D). Fig.~\ref{fig:vpr}E gives an example of a correct (top) and incorrect (bottom) match from $vision$, where the ground truth matched query and database reference image is shown, and the model predicted match. For the incorrect condition, the model aliased a place prediction that looked similar (Fig.~\ref{fig:vpr}E). 

\section*{Discussion}
The function of any visual system, natural or artificial, is to extract meaningful visual information.  Visual information is dense, and a key aspect of meaningful extraction is downsampling the information to a manageable volume and rate while enhancing the salience of the signal. This is especially true for animal brains in which processing is comparatively slow and metabolic costs seriously curtail computational resources.  Even among the animals, the constraints on visual system design are especially demanding for fast-flying and highly visual insects, including flies, bees and wasps, which must process images gathered at high rates from large, near panoramic eyes with tiny brains containing just hundreds of thousands of neurons in total~\cite{Chittka2009}.

We here presented a generalisable model (Fig.~\ref{fig:vizmodule}) of image processing in the insect visual system that achieved  effective downsampling of natural images while maintaining the distinguishing feature information in the images. Our model was capable of producing unique downsampled visual representations for both flowers and scenes (Fig.~\ref{fig:ann}-\ref{fig:snnselectivity}). Simple binary classification, as shown in Experiment 1, demonstrated a strong proof of concept that the proposed vision model produces sparse output activations of approximately 50 neurons that were distinct for and distinguished between stimuli. Experiment 2 tested the vision model further by evaluating if 17 different types of flowers could be distinguished with a sparse output code. We found that both our ANN and SNN versions of the vision model were capable of robustly discriminating between different flowers under varying implementations of inference. Finally, in Experiment 3 we showed that in a visual localization task that was dependent on discriminating between different scenes and recognising as similar scenes from the same location our vision models achieved better performance than a benchmark, simple image downsampling system (Fig.~\ref{fig:vpr}). The choice to use a spiking or non-spiking version of the vision model, therefore, strongly depends on the downstream use case and integration with additional models. 

Our model performed well in all three experiments with different image classes and datasets (Fig.~\ref{fig:ann}-\ref{fig:vpr}) with no manual tailoring of model parameters. This contrasts with previous models that were built bespoke for a particular problem class or image set and therefore had limited generalisability. For example, the neuromorphic vision model introduced by MaBouDi~\emph{et al.} excels at pattern recognition tasks~\cite{MaBouDi2025}, but only through training on the downstream test data. Our model highlights an important generalizability inherent through the self-supervised learning of images and scenes that are unrelated to the downstream test task, as shown through our Experiments 1-3.

We achieved this generalisability by using a series of convolutional neural nets that each aimed to approximate the kind of processing performed in each layer of insects' visual system, if not the computation by which the processing is implemented. In this regard, our model deviates from simulating the biological computations in insect visual processing as it is unlikely that biological brains learn by backpropagation and process through convolution. Additionally, it is also unclear to what degree there is any kind of experience-dependent plasticity in the layers of the insect visual processing system. 

However, we propose that the tuning we achieve in our model by convolution can be considered analogous to the optimisation of connectivity in the insect visual system that would be achieved across evolutionary and developmental timescales. It is notable that minimal self-supervised learning is needed to dramatically improve performance of the model. Importantly, and in contrast to other models~\cite{MaBouDi2025}, our experimental tasks did not require learning of the specific flowers or scenes to perform accurate classification. We propose that we are not teaching the vision model to learn how to recognize these things, but rather wire a neural network system is such a way that the diversity of visual input is robustly and reliably downsampled to a unique sparse code. We observe significant improvements in model performance with a single epoch of training, often without improvement during further training bouts, meaning that our sparse coding strategy learns quickly and becomes rigid through tuning of model weights.

The result is a model of visual processing and sparse neural coding that could be used as a "vision module" for any modelling of downstream learning, sensory integration, or decision making in insects. Additionally, our model can be integrated into neuromorphic artificial systems seeking to capture high levels of performance with limited compute. We have developed both ANN and SNN implementations of our model to maximise options for using out vision module in further modelling studies.  

\section*{Acknowledgments}

ADH~acknowledges continued support from the Queensland University of Technology (QUT) through the Centre for Robotics. This work was funded by the Australian Research Council (DP230100006, ABB and KN) and the US Air Force Office of Scientific Research (AFOSR, FA9550-23-1-0473, KN). ABB received support from the Macquarie University Bioinnovation Initiative.

\bibliography{plos_bibtex_sample}

\end{document}